\documentclass[letterpaper]{article} % DO NOT CHANGE THIS
\usepackage{aaai2026}  % DO NOT CHANGE THIS
\usepackage{times}  % DO NOT CHANGE THIS
\usepackage{helvet}  % DO NOT CHANGE THIS
\usepackage{courier}  % DO NOT CHANGE THIS
\usepackage[hyphens]{url}  % DO NOT CHANGE THIS
\usepackage{graphicx} % DO NOT CHANGE THIS
\usepackage{natbib}  % DO NOT CHANGE THIS AND DO NOT ADD ANY OPTIONS TO IT
\usepackage{caption} % DO NOT CHANGE THIS AND DO NOT ADD ANY OPTIONS TO IT
\usepackage{booktabs}
\usepackage{makecell}
\usepackage{amsmath}
\usepackage{algorithm}
\usepackage{algorithmic}

\usepackage{newfloat}
\usepackage{listings}
\usepackage{xcolor}
\DeclareCaptionStyle{ruled}{labelfont=normalfont,labelsep=colon,strut=off} % DO NOT CHANGE THIS
\floatstyle{ruled}
\newfloat{listing}{tb}{lst}{}
\floatname{listing}{Listing}
\title{A Fuzzy Logic Prompting Framework for Large Language Models in Adaptive and Uncertain Tasks}
\author {
    Vanessa Figueiredo
}
\affiliations{
    Department of Computer Science\\
    University of Regina\\
    vanessa.figueiredo@uregina.ca
}

\usepackage{bibentry}
\nocopyright

\begin{document}

\maketitle
\renewcommand{\thefootnote}{}
\footnotetext{Under review.}

\begin{abstract}
We introduce a modular prompting framework that supports safer and more adaptive use of large language models (LLMs) across dynamic, user-centered tasks. Grounded in human learning theory, particularly the Zone of Proximal Development (ZPD), our method combines a natural language boundary prompt with a control schema encoded with fuzzy scaffolding logic and adaptation rules. This architecture enables LLMs to modulate behavior in response to user state without requiring fine-tuning or external orchestration. In a simulated intelligent tutoring setting, the framework improves scaffolding quality, adaptivity, and instructional alignment across multiple models, outperforming standard prompting baselines. Evaluation is conducted using rubric-based LLM graders at scale. While initially developed for education, the framework has shown promise in other interaction-heavy domains, such as procedural content generation for games. Designed for safe deployment, it provides a reusable methodology for structuring interpretable, goal-aligned LLM behavior in uncertain or evolving contexts.
\end{abstract}

% Uncomment the following to link to your code, datasets, an extended version or similar.
% You must keep this block between (not within) the abstract and the main body of the paper.
% \begin{links}
%     \link{Code}{https://aaai.org/example/code}
%     \link{Datasets}{https://aaai.org/example/datasets}
%     \link{Extended version}{https://aaai.org/example/extended-version}
% \end{links}

\section{Introduction}
Imagine guiding a student who is just beginning to learn genetics. Rather than introducing complex terms like “epistasis,” an effective tutor identifies what the student already knows—perhaps they have heard of DNA—and builds on that foundation incrementally. This technique, known as instructional scaffolding, aligns support with the learner’s evolving capabilities and is formalized through the Zone of Proximal Development (ZPD) \cite{vygotsky_mind_1978}.

Large Language Models (LLMs) excel at general-purpose reasoning but struggle in user-adaptive or ambiguous scenarios, especially in \textit{fuzzy} domains like tutoring, where knowledge states are partial, rules are soft, and multiple valid responses may exist \cite{zadeh-1994-soft}. Without explicit structural guidance, LLMs often produce inconsistent, off-target, or contradictory outputs \cite{liu2023lost, gao-etal-2021-making}.

We propose a prompting framework inspired by cognitive scaffolding and fuzzy control. It consists of two components: (1) a natural language boundary prompt that defines the instructional role and tone, and (2) a control schema encoded with fuzzy scaffolding logic, which captures learner profiles, reasoning thresholds, and adaptation rules. This modular structure enables LLMs to handle ambiguity, adjust support levels, and maintain coherence without fine-tuning or external orchestration.

Our method bridges learning science, symbolic reasoning, and human tutoring strategies \cite{vygotsky_mind_1978, alma9922996293003476}, supporting safer and more interpretable use of LLMs in evolving instructional contexts. We evaluate the approach in a simulated tutoring setting using synthetic K–12 learner profiles. Models must modulate tone and strategy across tasks, grade levels, and prior knowledge bands. Though simulation-based, the setup reflects real classroom dynamics and serves as a pre-deployment feasibility study.

\textbf{Research Goals}
\begin{itemize}
\item Assess pedagogical alignment of LLM responses across varied learner states.
\item Compare scaffolding-based prompts to standard prompting baselines.
\item Analyze how prompt structure influences adaptivity and instructional quality.
\end{itemize}

\textbf{Contributions}
\begin{enumerate}
    \item A boundary-based prompting strategy grounded in cognitive learning theory.
    \item A hybrid prompt architecture that combines natural language with a control schema encoded with fuzzy scaffolding logic.
    \item Empirical results showing improved scaffolding, consistency, and relevance in tutoring tasks.
\end{enumerate}

While focused on education, the framework generalizes to other interaction-heavy domains such as simulation-based learning, personalized planning, and procedural content generation. It also supports emerging benchmarking efforts (e.g., TutorEval, LLM4Ed) that prioritize adaptivity and pedagogical fidelity.

\section{Related Work}

\subsection{Curriculum and Pedagogical Training}
Curriculum learning posits that sequencing tasks from simple to complex improves learning \cite{bengio2009-curriculum}. This principle has been applied to LLMs through curriculum-tuned training \cite{10.1145/3589335.3641257} and pedagogical prompting \cite{lee-from-2024}, where reasoning steps and difficulty progressions are embedded into prompts. However, these approaches focus on input design rather than real-time adaptation. In contrast, our method enables inference-time scaffolding using a control schema encoded with fuzzy scaffolding logic, aligned with learner state and instructional intent.

\subsection{Prompt Engineering}
Prompting strategies such as few-shot learning \cite{brown-2020-language}, chain-of-thought (CoT) prompting \cite{NEURIPS2022_9d560961}, and self-consistency decoding \cite{zhou2024teachingassistantintheloopimprovingknowledgedistillation} improve reasoning performance on structured tasks. However, these methods often assume static task structures and lack robustness in fuzzy or ill-defined scenarios. While personalized prompting has emerged \cite{lee-from-2024}, it typically relies on hardcoded templates \cite{peeters-entity-2024}. Our approach modularizes prompt design into declarative boundary instructions and a control schema with fuzzy scaffolding logic, supporting more dynamic adaptation and consistent behavior across learner profiles.

\subsection{Agent Framing for LLMs}
Framing LLMs as agents with roles, goals, and beliefs improves coherence, alignment, and goal-driven outputs \cite{andreas-2022-language}. We build on this perspective by assigning the model an instructional role and regulating its behavior via a control schema, guiding the system toward pedagogically consistent outputs in interactive scenarios.

\subsection{Structured Prompting with Control Schemas}
Recent work has shown that LLMs can parse and follow structured inputs, such as JSON, more reliably in multi-step domains like planning and robotics \cite{okuda-askit-2024, El-Teleity-2011-fuzzy}. We adapt this insight by embedding a control schema encoded with fuzzy scaffolding logic, specifying task types, learner knowledge levels, and adaptive strategies. This structured representation promotes interpretability, facilitates modular updates, and constrains model behavior without exhaustive prompt engineering.

\subsection{Fuzzy Logic and Symbolic Control}
Fuzzy logic models uncertainty using soft constraints and linguistic variables \cite{zadeh-1994-soft, MendelJ.M.1995Flsf}. Fuzzy control has been applied in domains such as dialogue management and educational planning to support ambiguity-tolerant reasoning \cite{griol-2021-adaptive, Abduljabbar-2023-high}. Although LLMs are inherently probabilistic, they lack native mechanisms for fuzzy control. Our framework introduces an external control schema that encodes fuzzy scaffolding logic, serving as a symbolic controller that supports nuanced, rule-aligned behavior. This approach also avoids context degradation common in long in-context learning chains \cite{gao-etal-2021-making}, preserving instructional fidelity through a declarative, reusable structure.

\subsection{Neural-Symbolic Prompting}
Neural-symbolic systems aim to combine the expressive power of neural models with the interpretability and rule-based control of symbolic logic \cite{garcez2019neuralsymboliccomputingeffectivemethodology}. While our models remain purely neural, we follow the neural-symbolic prompting paradigm by externalizing rule encoding into a control schema with fuzzy scaffolding logic. This lightweight symbolic interface guides inference without modifying the model architecture.

\subsection{LLMs in Education}
LLMs are increasingly used in educational applications, including intelligent tutoring systems (ITSs), automated feedback, and assessment tools \cite{10.1145/3568812.3603476, 10260740}. However, many rely on static templates, curated responses, or human-in-the-loop supervision. Our framework enables real-time, adaptive instructional support by structuring prompts through a boundary layer and an external control schema encoded with fuzzy scaffolding logic, offering more scalable, interpretable personalization in educational contexts.

\section{Problem Definition}
Despite their broad reasoning abilities, LLMs struggle in tasks that are underspecified, user-dependent, or dynamically evolving, hallmarks of \textit{fuzzy logic} environments. This is especially critical in domains like personalized tutoring, coaching, and interactive simulations, where ideal responses depend on partial user knowledge, shifting goals, and soft constraints.

Standard prompting methods, including few-shot learning, instruction tuning, and CoT, assume fixed task formats and are constrained in real-time adaptation. As a result, LLMs often produce outputs that are overly generic, misaligned, or inconsistent across turns. Moreover, while LLMs are probabilistic, they lack native support for reasoning with fuzzy rules or maintaining behavioral coherence in uncertain contexts.

We define the problem as follows:

\textit{Given a context with fuzzy constraints, such as evolving learner knowledge, ambiguous inputs, or partial signals, how can we design a prompting strategy that allows an LLM to:}
\begin{enumerate}
    \item Interpret user state along a continuum of understanding,
    \item Adjust tone, support level, and pedagogical strategy accordingly, and
    \item Maintain coherence across multi-turn interactions.
\end{enumerate}

Solving this requires bridging flexible language generation with structured guidance. Our approach enables LLMs to act as adaptive agents, modulating behavior in response to contextual cues without fine-tuning or external tooling.

To achieve this, we introduce a modular prompting framework that separates stable declarative intent (e.g., instructional role and tone) from dynamic task logic (e.g., knowledge state, fallback strategies). This separation is implemented through a natural language boundary prompt and a JSON encoded with fuzzy scaffolding logic, enabling adaptive behavior in fuzzy environments through symbolic prompting.

\section{Scaffolding Logic}
\subsection{Overview}
Fuzzy logic enables reasoning under uncertainty through graded rules, while the Zone of Proximal Development (ZPD) models learning as progression across a continuum of understanding with targeted support. Our framework integrates these ideas by mapping ZPD-inspired scaffolding bands, such as \textit{emerging}, \textit{developing}, and \textit{proficient}, to fuzzy control states within a structured schema.

This alignment allows LLMs to interpret learner signals as continuous rather than categorical, adjusting responses based on confidence and support needs. The result is a neural-symbolic prompting framework that encodes soft behavioral boundaries and enables context-aware adaptivity in instructional and other dynamic tasks.

\begin{figure}[t]
\centering
\includegraphics[width=0.9\columnwidth]{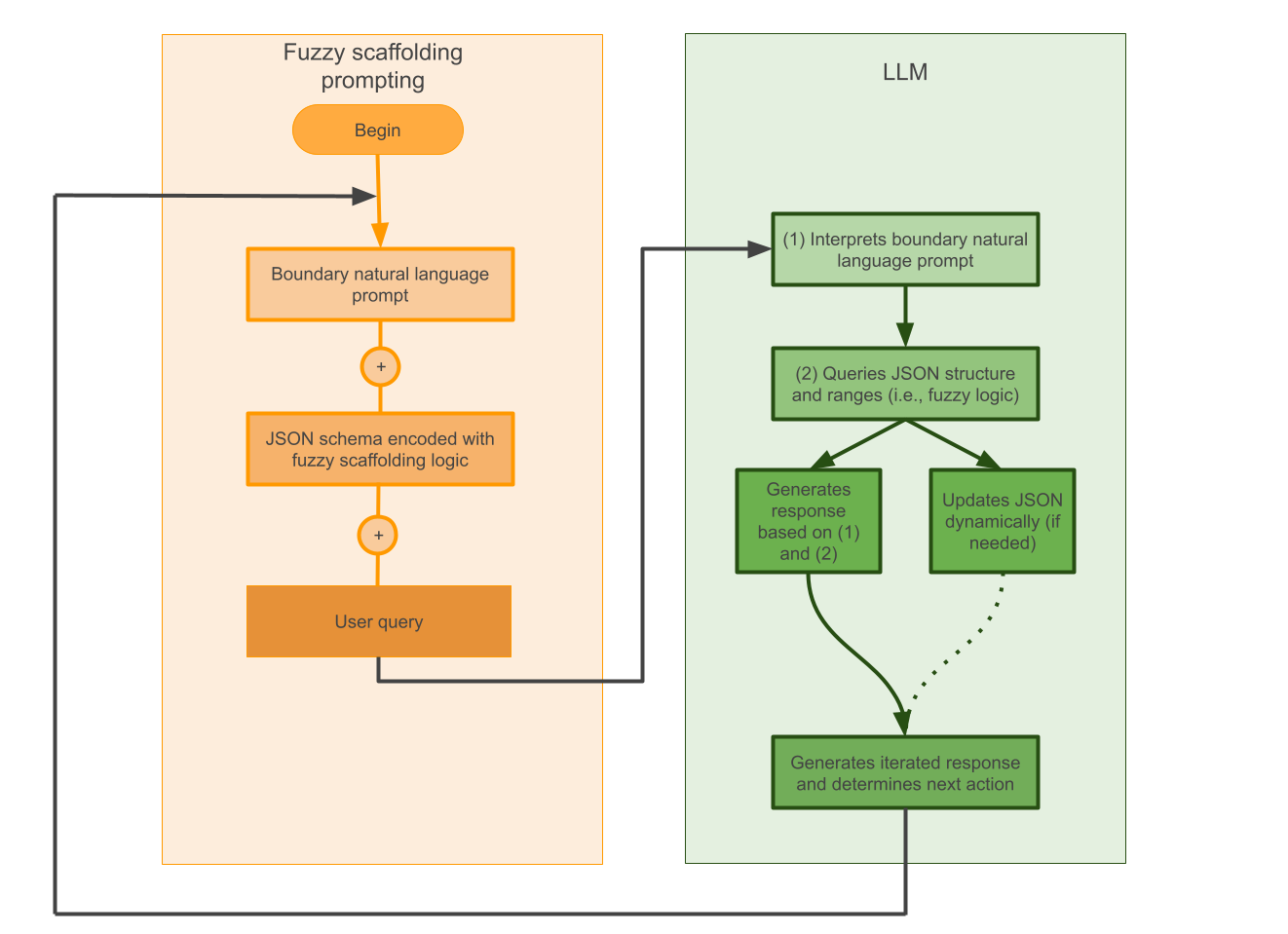}
\caption{Fuzzy Scaffolding Inference Workflow. The boundary prompt and JSON schema guide LLM behavior, enabling context-sensitive scaffolding through fuzzy rule lookup and adaptive generation.}
\label{fig1}
\end{figure}

As shown in Figure~\ref{fig1}, our method introduces a two-layer prompt design:

\begin{enumerate}
    \item \textbf{Boundary Prompt (Outer Layer)}: A natural language description of system role, tone, task domain, and pedagogical goals.
    \item \textbf{JSON Schema encoded with fuzzy scaffolding logic (Inner Layer)}: A structured object that encodes task logic, fuzzy support states, and learner-specific contingencies.
\end{enumerate}

\subsection{Boundary Prompt}
The boundary prompt establishes the instructional setting and links to the scaffold. For example:

\begin{lstlisting}
You are a tutor that adapts to a student's grade, task, and knowledge level.

Use `scaffolding_recipe.json` to:
1. Match the task type.
2. Apply a strategy based on knowledge level.
3. Choose scaffolding support.
4. Adjust vocabulary to grade level.
5. Monitor learning using update rules.

Always ask if the student understood before moving on.
\end{lstlisting}

This defines the model’s context, scope, and adaptation policy.

\subsection{Fuzzy Scaffolding logic JSON}
The scaffold provides a declarative specification of fuzzy instructional logic. It includes:

\begin{itemize}
    \item Task types: e.g., recall, comprehension, computation.
    \item ZPD bands: \verb|emerging|, \verb|developing|, etc.
    \item Support levels: aligned to scaffolding needs (e.g., high support → task breakdown).
    \item Readability targets: based on grade level (e.g., Flesch-Kincaid).
    \item Adaptation rules: e.g., how to update support based on learner progress.
\end{itemize}

\begin{lstlisting}
"knowledge_levels": {
  "emerging": {
    "description": "Low prior knowledge. Triggered by phrases like 'I'm not sure'."
  }
},
"scaffolding_types": {
  "high": {
    "description": "Break down tasks, provide guided examples."
  }
}
\end{lstlisting}

This schema is referenced at runtime to guide behavior based on user signals, without overloading the prompt.

\subsection{Execution Flow}
At inference time, the model executes the following steps:

\begin{enumerate}
    \item Parse user profile and context via the boundary prompt.
    \item Query the JSON scaffold to:
        \begin{itemize}
            \item Determine task type and support level.
            \item Adapt tone and vocabulary by grade.
            \item Select appropriate follow-up (hint, rephrase, etc.).
        \end{itemize}
    \item Generate an adaptive response.
    \item Offer continuation options based on scaffold logic.
\end{enumerate}

This design supports flexible, interpretable, and pedagogically consistent interaction, enabling symbolic control without modifying the model or requiring external agents.

\section{Experimental Setup}
We evaluated our scaffolded modular prompting framework (natural language boundary prompt + fuzzy scaffolding logic JSON) in a simulated intelligent tutoring system (ITS) for middle school students. The study addresses three research goals:
\begin{itemize}
    \item \textbf{Goal 1}: Assess the pedagogical quality of LLM-generated tutoring responses across learner profiles.
    \item \textbf{Goal 2}: Compare our modular prompting to standard prompting strategies.
    \item \textbf{Goal 3}: Examine how prompt structure affects adaptivity, scaffolding quality, and readability across tasks and grade levels.
\end{itemize}

\subsection{Models Evaluated}
We tested four models:
\begin{itemize}
    \item GPT-3.5 Turbo (OpenAI API): temperature = 0.7, max tokens = 300.
    \item Gemma 3 4B, LLaMA 3.1 8B, and DeepSeek-R1 8B: deployed via Ollama with default generation settings.
\end{itemize}

\subsection{Prompting Conditions}
Each model was evaluated under four prompting strategies:
\begin{enumerate}
    \item Scaffolded Modular Prompt (Ours): boundary instruction + JSON schema.
    \item Flat Prompt: "You are a helpful assistant."
    \item CoT: "You are a tutor. Think step by step" \cite{kojima-2022-large}.
    \item Few-Shot Prompt: Based on Brown et al. (\citeyear{brown-2020-language}), with exemplar-based tutoring demonstrations.
\end{enumerate}

\subsection{Synthetic Data Generation}
We created 200 synthetic tutoring scenarios using GPT-4: 100 math and 100 science prompts across Grades 6 and 8. Each scenario included:
\begin{itemize}
    \item A subject-specific task (e.g., computation, reasoning, recall).
    \item A student profile with one of four scaffold levels: \textit{emerging}, \textit{developing}, \textit{proficient}, or \textit{advanced}.
\end{itemize}

Each interaction included three turns to assess dynamic adaptation.

\subsection{Evaluation Metrics}
We used rubric-based evaluation to assess pedagogical effectiveness. A GPT-4 evaluator rated each response on a 5-point Likert scale across three dimensions:
\begin{enumerate}
    \item Grade Appropriateness
    \item Scaffolding Quality
    \item Adaptivity Across Turns
\end{enumerate}

Inputs to the evaluator included the student profile, prompt, and model response. LLM-based assessment provided consistency and scale, functioning as a practical substitute for expert human annotation at this exploratory stage.

\subsection{Implementation}
A Python pipeline simulated tutoring sessions and logged model responses, LLM ratings, and NLP metadata into structured CSVs. All models were tested under identical conditions using fixed prompts and deterministic settings for comparability.

Code, prompts, and evaluation outputs will be released upon acceptance to support transparency and reproducibility.

\section{Results}

We evaluated model performance across three dimensions: \textit{grade-level alignment}, \textit{scaffolding quality}, and \textit{adaptivity}. These were assessed using rubric-based LLM evaluations across four prompting strategies: \textit{flat}, \textit{few-shot}, \textit{CoT}, and our \textit{fuzzy scaffolding logic} (recipe). We also performed an ablation study to isolate the impact of prompt structure.

\subsection{Rubric-Based Evaluation}

This analysis addresses \textbf{Goals 1} and \textbf{2}, showing that both model type and prompt structure significantly affect educational response quality.

\begin{figure}[ht]
\centering
\includegraphics[width=0.48\textwidth]{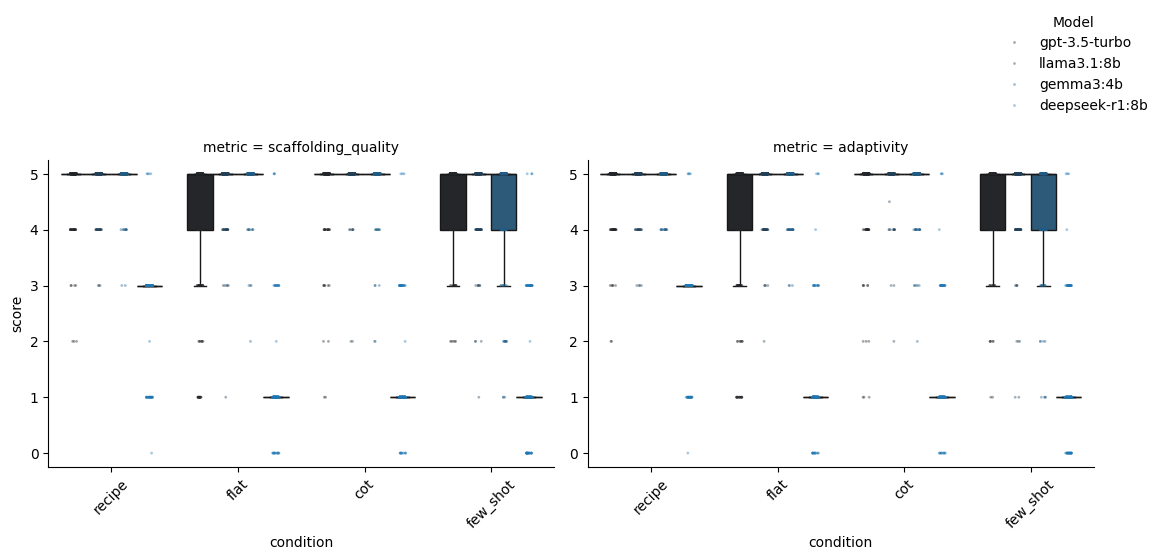}
\caption{Rubric scores across prompting strategies. \textit{Recipe} refers to scaffolded prompts. Red lines indicate \( \text{mean} \pm \text{SEM} \).}
\label{fig:facet_boxplots}
\end{figure}

\textbf{Key Findings:}
\begin{itemize}
    \item \textbf{Grade Match:} Scaffolded prompts (recipe) outperformed others (M=4.42), significantly surpassing flat and few-shot prompts (\(p<.001\)). Though Kruskal-Wallis tests yielded weak global effects (\(p=0.48\)), pairwise differences were consistent across models.
    
    \item \textbf{Scaffolding Quality:} Scaffolded prompts (recipe) led to significantly higher scores vs. all baselines (\(p<.001\)), with moderate effect sizes (e.g., \(d=0.45\), \(\delta=0.19\) vs. few-shot).
    
    \item \textbf{Adaptivity:} Similar trends were observed, with scaffolded prompts (recipe) achieving the highest scores (\(p<.001\), \(d=0.46\), \(\delta=0.20\)).
\end{itemize}

\begin{figure}[ht]
\centering
\includegraphics[width=0.48\textwidth]{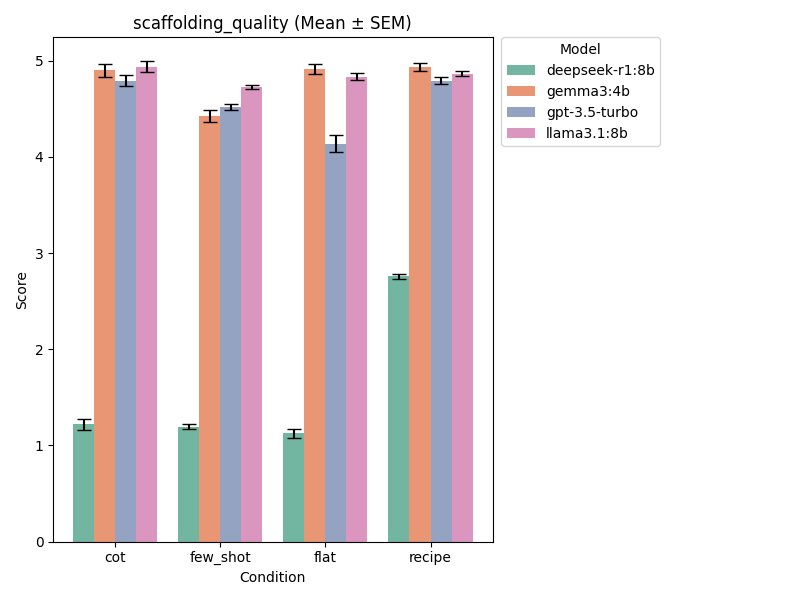}
\includegraphics[width=0.48\textwidth]{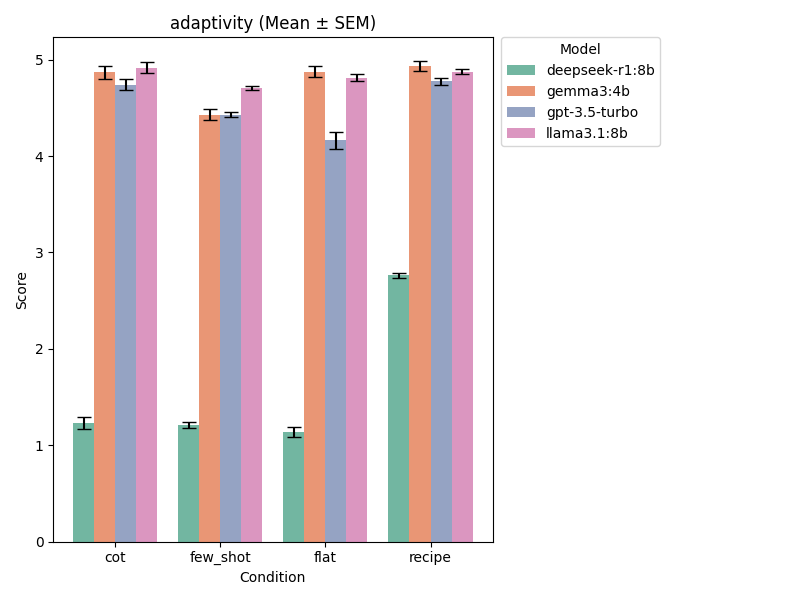}
\caption{Mean rubric scores (\( \pm \text{SEM} \)) by prompt type. Scaffolded prompts (recipe) consistently outperform others.}
\label{fig:barplots_sem}
\end{figure}

\begin{figure}[ht]
\centering
\includegraphics[width=\linewidth]{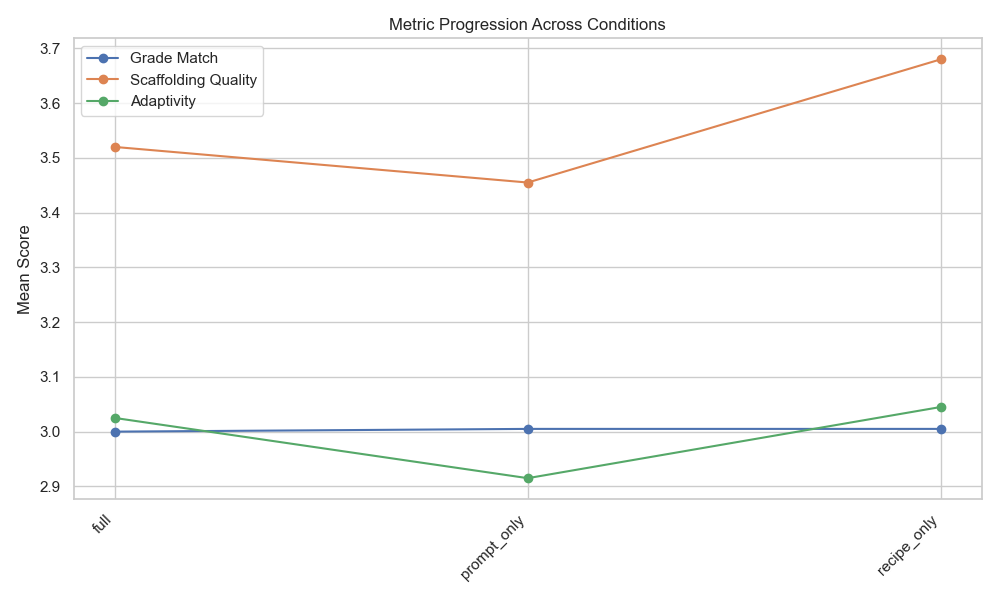}
\caption{Progression across dialogue turns. Scaffolded prompts (full) yield stronger gains.}
\label{fig:lineplot_progression}
\end{figure}

\begin{table}[ht]
\small
\centering
\caption{Statistical comparison of rubric scores. Scaffolded prompts (SP) show consistent improvements.}
\label{tab:rubric_stats}
\begin{tabular}{@{}p{1.3cm}p{2.2cm}ccccc@{}}
\toprule
Rubric & Comparison & Kruskal $p$ & $\eta^2$ & $d$ & $\delta$ \\
\midrule
Grade & SP vs Flat & 0.485 & 0.01 & 0.24 & 0.03 \\
Grade & SP vs Few-shot & 0.485 & 0.01 & 0.22 & 0.03 \\
Scaffold & SP vs Flat & $<$0.001 & 0.02 & 0.41 & 0.13 \\
Scaffold & SP vs Few-shot & $<$0.001 & 0.02 & 0.45 & 0.19 \\
Adaptivity & SP vs Flat & $<$0.001 & 0.03 & 0.41 & 0.14 \\
Adaptivity & SP vs Few-shot & $<$0.001 & 0.03 & 0.46 & 0.20 \\
\bottomrule
\end{tabular}
\end{table}

\vspace{-1em}

\subsection{Ablation Study}
To assess the contribution of each component in our prompt design (\textbf{Goal 3}), we compared:
\begin{itemize}
    \item \textbf{Full}: boundary prompt + JSON encoded with fuzzy scaffolding logic
    \item \textbf{Prompt-only}: boundary prompt without JSON encoded with fuzzy scaffolding logic
    \item \textbf{Scaffold-only}: JSON encoded with fuzzy scaffolding logic without boundary prompt
\end{itemize}

\begin{figure}[ht]
\centering
\includegraphics[width=\linewidth]{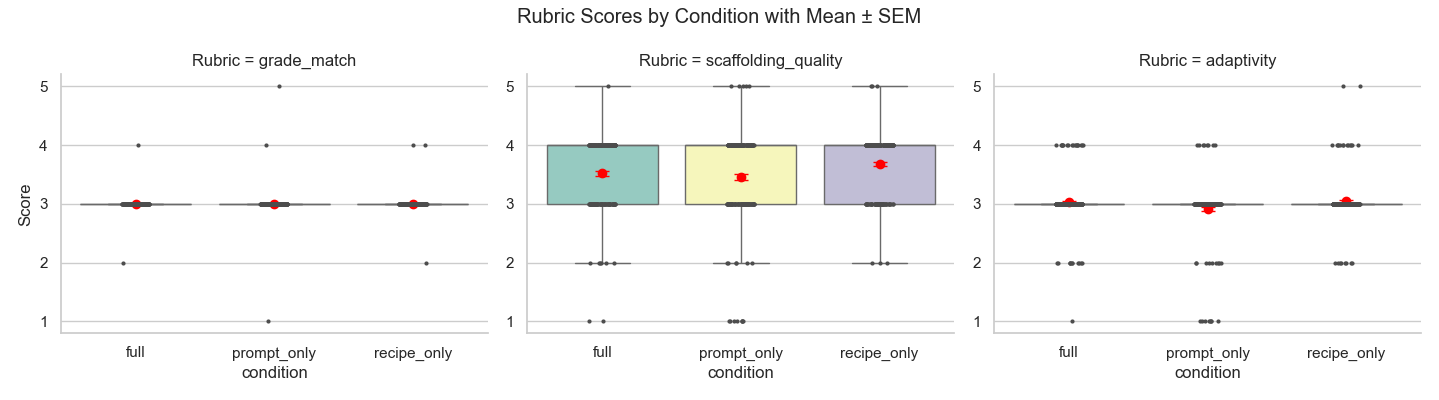}
\caption{Ablation results by condition. Scaffold-only performed best on scaffolding quality.}
\label{fig:ablation_facet}
\end{figure}

\textbf{Findings:}
\begin{itemize}
    \item \textbf{Grade Match}: No significant differences (\(p=0.78\)).
    \item \textbf{Scaffolding Quality}: Scaffold-only outperformed prompt-only (\(p=0.03\), \(d=1.06\), \(\delta=0.67\)), highlighting the instructional value of the schema.
    \item \textbf{Adaptivity}: No significant effects, but moderate effect size for Full vs. Prompt-only (\(d=0.91\), \(\delta=0.39\)).
\end{itemize}

\begin{table}[ht]
\centering
\caption{Ablation test statistics. Scaffold component drives most gains.}
\label{tab:ablation_stats}
\resizebox{\linewidth}{!}{%
\begin{tabular}{llcccc}
\toprule
Rubric & Comparison & Friedman $p$ & Dunn $p$ & $d$ & $\delta$ \\
\midrule
Grade & Full vs Prompt-only & 0.779 & 1.000 & -0.18 & -0.14 \\
Scaffold & Scaffold vs Prompt-only & 0.084 & 0.009 & 1.06 & 0.67 \\
Scaffold & Full vs Scaffold & 0.084 & 0.026 & 0.92 & 0.56 \\
Adaptivity & Full vs Prompt-only & 0.311 & 0.067 & 0.91 & 0.39 \\
\bottomrule
\end{tabular}%
}
\end{table}

\textbf{Interpretation:} Gains are not due to verbosity or prompt length, but to the structured interplay between fuzzy scaffolding logic and boundary prompt.

All distributions violated normality and homogeneity assumptions (Shapiro-Wilk and Levene \(p<.001\)); we report both non-parametric (Kruskal, Dunn) and effect size metrics for robustness.

\section{Discussion}
Our findings demonstrate that scaffolded prompts, combining a boundary prompt with a control schema encoded with fuzzy scaffolding logic, substantially improve adaptivity and instructional alignment in LLM-based ITSs. By structuring task logic and support rules declaratively, this framework enables LLMs to reason under soft constraints, similar to classical fuzzy control systems \cite{zadeh-1994-soft}. In contrast to prior approaches that rely on lengthy demonstrations or static templates \cite{gao-etal-2021-making, peeters-entity-2024}, our method introduces a lightweight, interpretable structure that reduces long-context degradation while supporting dynamic behavioral modulation.

Framing the LLM as a pedagogical agent via the boundary prompt aligns with the agent modeling perspective \cite{andreas-2022-language}, encouraging consistent, goal-directed behavior across evolving learner interactions. Though our evaluation was conducted in a synthetic simulation, the architecture is model-agnostic and has shown early promise in adjacent domains such as procedural content generation for games and assistive dialogue systems.

Limitations include the constrained subject and grade coverage, as well as the use of synthetic tasks rather than real student interactions. Future work will extend this framework to live classroom settings, teacher-in-the-loop adaptations, and cross-domain applications. Importantly, the separation of declarative context from adaptive control provides a scalable path toward interpretable, flexible LLM behavior in settings marked by ambiguity or evolving user needs.

\subsection{Toward benchmarks for adaptive LLM tutors}
In addition to prompting evaluation, this work supports the development of structured benchmarks for educational LLM applications. Our task set, scoring rubric, and LLM-assisted evaluation pipeline align with emerging initiatives such as \textit{TutorEval}, which prioritize pedagogical alignment, scaffolding quality, and learner adaptivity over static QA performance. By operationalizing fuzzy scaffolding logic within a reusable control schema, this study contributes a modular and extensible framework for future benchmarking efforts—laying the groundwork for a TutorEval suite that includes real student data, domain-specific challenges, and human-in-the-loop evaluation.

\section{Conclusion}
We presented a scaffolded prompting framework that enhances the adaptivity and instructional quality of LLMs by drawing on principles from cognitive learning theory and fuzzy logic. The framework consists of two modular components: a boundary prompt that defines the system’s instructional role, tone, and domain constraints using natural language; and a control schema, encoded as a JSON object, that formalizes fuzzy scaffolding logic through declarative task rules, support strategies, and learner-state mappings.

This design bridges symbolic reasoning and statistical generation by operationalizing the Zone of Proximal Development (ZPD) as a fuzzy control system. Instead of relying on static rules or handcrafted demonstrations, our method treats instructional support as a graded continuum, enabling LLMs to dynamically adjust their behavior across ambiguous or evolving contexts. The structured prompt architecture provides both interpretability and adaptability—key properties for safety-critical or user-sensitive applications.

While this study focused on simulated intelligent tutoring, the same framework has been successfully extended to other interactive domains such as dialogue-based games and procedural content generation. This suggests broad applicability for scaffolded prompting in any setting that requires adaptive, user-centered reasoning, spanning education, training, simulation, and beyond. By uniting human learning strategies with symbolic prompt control, our work contributes a scalable methodology for building more responsive, pedagogically grounded AI systems.

\bibliography{sample}

\end{document}